\pdfoutput=1
\documentclass[11pt]{article}

\usepackage{acl}

\usepackage{times}
\usepackage{latexsym}
\usepackage{multirow}

\usepackage[T1]{fontenc}
\usepackage[utf8]{inputenc}

\usepackage{microtype}

\usepackage{eso-pic} 
\usepackage{algorithm}
\usepackage{algpseudocode}
\usepackage{amsmath}
\usepackage{amssymb}
\usepackage{url}
\usepackage{epsfig}
\usepackage{graphicx}
\usepackage{caption}
\usepackage{subcaption}

\title{Emotion-Cause Pair Extraction as Question Answering}

\author{Huu-Hiep Nguyen \\
  Cinnamon AI \\
  \texttt{hubert@cinnamon.is} \\\And
  Minh-Tien Nguyen \thanks{Corresponding author.}\\
  Hung Yen University of Technology and Education \\
  \texttt{tiennm@utehy.edu.vn} \\}

\begin{document}
\maketitle

\begin{abstract}
The task of Emotion-Cause Pair Extraction (ECPE) aims to extract all potential emotion-cause pairs of a document without any annotation of emotion or cause clauses. Previous approaches on ECPE have tried to improve conventional two-step processing schemes by using complex architectures for modeling emotion-cause interaction. In this paper, we cast the ECPE task to the question answering (QA) problem and propose simple yet effective BERT-based solutions to tackle it. Given a document, our \textit{Guided-QA} model first predicts the best emotion clause using a fixed question. Then the predicted emotion is used as a question to predict the most potential cause for the emotion. We evaluate our model on a standard ECPE corpus. The experimental results show that despite its simplicity, our Guided-QA achieves promising results and is easy to reproduce. The code of Guided-QA is also provided.
\end{abstract}

\section{Introduction}
\label{sec:intro}

\begin{figure}[h!]
\centering
\epsfig{file=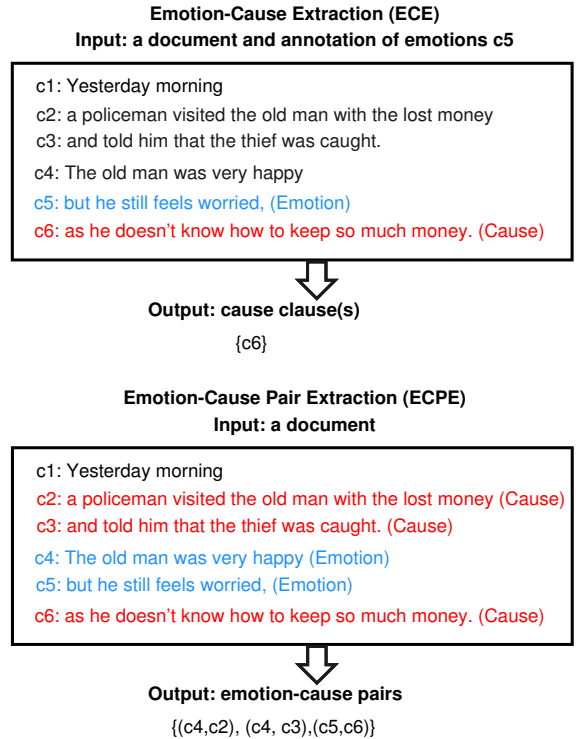, height=4.0in}
\caption{Illustration of ECE and ECPE tasks.}\vspace{0.0cm}
\label{fig:ecpe}
\end{figure}

%The ECE task was first proposed by \cite{lee2010text} and defined as a word-level sequence labeling problem.
Emotion Cause Extraction (ECE) is the task of detecting the cause behind an emotion given the emotion annotation \cite{lee2010text,gui2016event}, see Figure \ref{fig:ecpe} (Top). The text was divided into clauses and the task was to detect the clause containing the cause, given the clause containing the emotion. However, the applicability of ECE is limited due to the fact that emotion annotations are required at test time. Recently, \cite{xia2019emotion} introduced the more challenging Emotion-Cause Pair Extraction (ECPE) task: extracting all possible emotion-cause clause pairs in a document without annotations. Figure \ref{fig:ecpe} (Bottom) shows an example of the ECPE task. The input is a document of six clauses. Clauses c4 and c5 contain emotion with the emotion expressions “happy” and "worried". The emotion c4 has two causes c3 and c2, the emotion c5 has one cause c6, so the expected output is \{(c4,c2), (c4,c3), (c5,c6)\}.
%The extraction is usually done by extracting emotion-cause pairs \cite{ding2020end,wei2020effective,chen2020end,yan2021position}.
%We use the following toy example to illustrate the point. 

% The ECPE task is also important for root-cause analysis \cite{sole2017survey, hassanzadeh2019answering}. Automatic and accurate extraction of potential cause and effect pairs is critical for building pipelines of causal analysis. In this context, the effect is equivalent to emotion of the ECPE task. The better cause-effect pair extraction is, the more reliable subsequent processing modules are.

%% Toy example "Why emotion-cause pair?
\textit{Why cause-effect pair extraction?} 
We argue that \textit{independent} extraction of cause and emotion may be ineffective. For a given document, ECPE models may predict correct cause but incorrect emotion. This makes the output \textit{incomplete}, and subsequent processing steps less reliable \cite{ding2020end,wei2020effective,chen2020end,yan2021position}. We make a toy example of two models using the document in Figure \ref{fig:ecpe}. Model-1 predicts (c4,c1) and (c6,c3) as emotion-cause pairs. Its emotion, cause and pair accuracy scores are 0.5, 0.33 and 0.0. Model-2 predicts (c4, c2) and (c6, c1) as emotion-cause pairs. Its emotion, cause and pair accuracy scores are 0.5, 0.33 and 0.33. From the perspective of the pair extraction task, Model-2 is better.

%Question Answering (QA) is a versatile task in NLP that accepts a variety of formats \cite{khashabi2020unifiedqa}. The task of clause-level emotion-cause extraction (ECE) is closely related to multiple-choice QA in which each clause is a choice. 
% such as extractive span selection, abstractive answer generative, multiple choice, yes-no answer etc. 
%Referring to Figure \ref{fig:ecpe}, if we know the emotion c4, we may formulate it as question, e.g. "why the old man is very happy?", then QA models try to detect the cause as answer. 

%% OLD content (see 3.4)
%Existing schemes such as \cite{ding2020end, wei2020effective, chen2020end} attempt to reduce the $O(n^2)$ complexity of emotion-cause pair classification by using sliding window, transition graph etc. However, these techniques may miss some interaction between the emotion-cause pair and the full context (document). BERT-based QA models with full attention between question and context mitigate this issue. Through QA, the emotion-cause relationship is implicitly modeled and we can leverage the power of existing QA frameworks.

%In this paper, we make use of BERT-family models \cite{devlin2018bert, liu2019roberta} and cast the ECPE task as extractive QA in SQuAD format \cite{rajpurkar2016squad}. Given a document, our proposed Guided-QA model first predicts the best emotion clause using a fixed question (e.g. "emotion"). Then the predicted emotion is used as question to predict the most potential cause for it. 

Previous studies addressed the ECPE task by using sequence labeling \cite{lee2010text,cheng2021unified}, clause-level classification \cite{gui2016event,ding2020end,chen2020end}, ranking \cite{wei2020effective}, or recurrent synchronization \cite{chen2022recurrent}. The methods achieved promising results, yet the use of interaction between emotion and cause clauses is still an open question. For example, c4 and c2 share "the old man" tokens, which refer to "him" in c3; and c5 and c6 share "he", which mentions "the old man" in c2 and c4. %Also, emotion clauses usually include emotion tokens (i.e., "very happy" or "feels worried"), which are beneficial for the reference of causes.

Based on this observation, we introduce a paradigm shift \cite{sun2022paradigm} for ECPE by using \textit{span extraction}. As far as we know, \cite{gui2017question} is the first work that uses question answering for emotion-cause detection. However, their work addresses the ECE task only, which requires the annotation of emotion for cause prediction. In contrast, our paradigm shift is applied to the ECPE task, which is more challenging and does not require the annotation of emotion for cause prediction. The paradigm bases on two hypotheses. First, information from emotion clauses can be used to infer cause clauses. Second, emotion and cause clauses share implicit interaction. The design of our model is based on these two hypotheses. For the first hypothesis, we form questions based on emotional information which is used to predict emotion clauses. For the second hypothesis, we used predicted emotion as the guided question for cause prediction. The model is trained by using the BERT-QA architecture \cite{devlin2018bert} in form of SQuAD task \cite{rajpurkar2016squad}.

%and a multi-hop memory network is used to model the interaction between emotion tokens and all tokens in a potential cause clause.

Our paper makes three main contributions. 
\begin{itemize}
    \item We formulate the ECPE task as a QA problem and propose a Guided-QA model to implicitly capture the relationship between emotion and cause clauses, in which the predicted emotion is used as a guided question for cause prediction. The model can capture the implicit interaction between emotions and causes with a simple but effective architecture. To the best of our knowledge, we are the first to address the ECPE task by using QA formulation.
    \item We evaluate our model on the standard ECPE corpus \cite{xia2019emotion,fan2020transition}. Experimental results show that our approach achieves promising results compared to previous methods.
    \item We promote the reproducibility \cite{houghton2020guaranteeing} by providing the source code of our methods as well as rerunning publicly available source codes of the compared methods.
\end{itemize}

% Our ablation study confirms the importance of the QA-based model for the whole iterative extraction of emotion-cause pairs.

\section {Related Work}
% ECE and ECPE
\paragraph{ECE and ECPE tasks}
%It has been extensively studied for years \cite{lee2010text,gui2016event}.
The ECE task was formulated as sequence-labeling by \cite{lee2010text} and refined as clause-level by \cite{gui2016event}. Recently, the more challenging ECPE task \cite{xia2019emotion} has attracted a lot of contributions %\footnote{https://github.com/stevehamwu/Emotion-Cause-Analysis-Papers} 
with several strong methods \cite{ding2020end,wei2020effective,chen2020end,cheng2021unified,chen2022recurrent}. For example, \cite{ding2020end} introduced ECPE-MLL, which uses a sliding window for a multi-label learning scheme. ECPE-MLL extracts the emotion and cause by using the iterative synchronized multitask learning. \cite{chen2022recurrent} proposed a similar approach, recurrent synchronization network (RSN), that explicitly models the interaction among different tasks.
\cite{wei2020effective} presented RankCP, a transition-based framework, by transforming the ECPE problem into directed graph construction, from which emotions and the corresponding causes can be extracted simultaneously based on labeled edges.  
The PairGCN model \cite{chen2020end} used Graph Convolutional Networks to model three types of dependency relations among local neighborhood candidate pairs and facilitate the extraction of pair-level contextual information. 

We share the purpose of addressing the ECE and ECPE tasks with prior studies, however, instead of using classification or sequence labeling, we address the tasks with a new paradigm shift by using span extraction. It allows us to take into account the implicit interaction between emotion and cause clauses and to design a simple but effective BERT-based model for ECE and ECPE.

\cite{bi2020ecsp} derived a span-based dataset and formulated a new  ECSP (Emotion Cause Span Prediction) task from \cite{xia2019emotion} but it has not attracted much attention. The accessibility of the dataset and source code may be the reason. We leave span-based ECSP evaluation as future work. 

\paragraph{Paradigm shift in natural language processing}
A paradigm is a general modeling framework or a family of methods to solve a class of tasks. For instance, sequence labeling is a mainstream paradigm for Part-of-speech (POS) tagging and Named entity recognition (NER). The sequence-to-sequence (Seq2Seq) paradigm is a popular tool for summarization and machine translation. Different paradigms usually require different formats of input and output, and therefore highly depend on the annotation of the tasks.

Paradigm shift indicates the job of solving one NLP task in a new paradigm by reformulating the task along with changing the input-output formats. Paradigm shift in NLP has been explored scatterringly in recent years and with the advent of pretrained language models, it became a rising trend \cite{li2019unified,khashabi2020unifiedqa}. An excellent survey of paradigm shifts in NLP has been done by \cite{sun2022paradigm}. In this work, we realize such a paradigm shift for the ECPE task, i.e., we reformulate the clause-based text classification task as span extraction.

% BERT-QA
\paragraph{Span-based extractive question answering}
Our formulation for the tasks of ECE and ECPE relates to span-based extractive QA, which has been widely investigated \cite{khashabi2020unifiedqa}. More precisely, we design our model based on the pretrained language models (PLMs) such as BERT \cite{devlin2018bert} or RoBERTa \cite{liu2019roberta}. This is because applying PLMs as the backbone of QA systems has become a standard procedure.
For detailed information, please refer to \cite{devlin2018bert}.

\begin{figure}
\centering
\epsfig{file=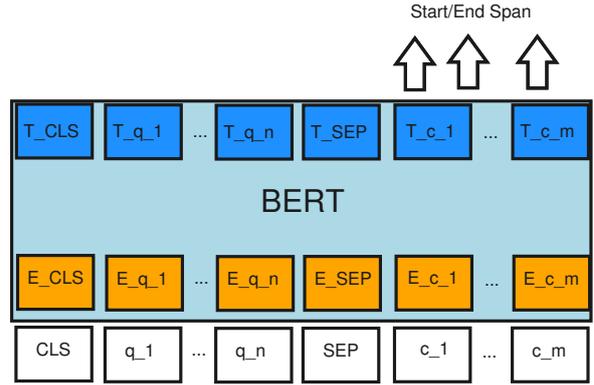, height=2.0in}
\setlength{\abovecaptionskip}{5pt}
\caption{BERT-based extractive Question Answering}
\label{fig:bert-qa}
\end{figure}

Figure \ref{fig:bert-qa} reproduced from \cite{devlin2018bert} shows how BERT is applied to the extractive QA task. Tokens of question $q = q_1,..,q_n$ and context $C=c_1,..,c_m$ are concatenated before being encoded by BERT. The contextual representations of tokens $T_i$ are put into a feed-forward layer followed by a softmax. Each candidate span for the answer is scored as the product of start/end probabilities. The maximum scoring span is used as the prediction. The training objective is the loglikelihood of the correct start and end positions.

By casting the ECPE to QA problem, our work leverages the powerful models of the BERT family \cite{devlin2018bert} to detect clause-level emotions and causes as well as emotion-cause pairs.

\section {Method}

\begin{figure*}[!t]
\centering
\epsfig{file=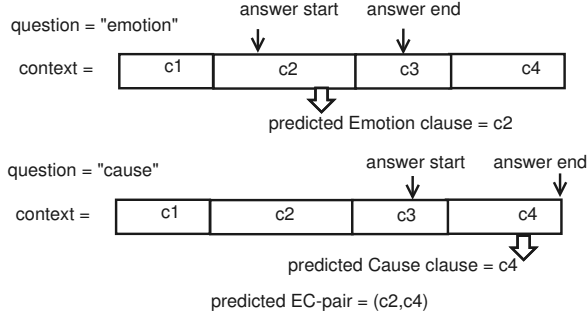, height=1.65in}
\caption{Independent extraction Indep-QA.}
\label{fig:indep-qa}
\end{figure*}

\begin{figure*}[!t]
\centering
\epsfig{file=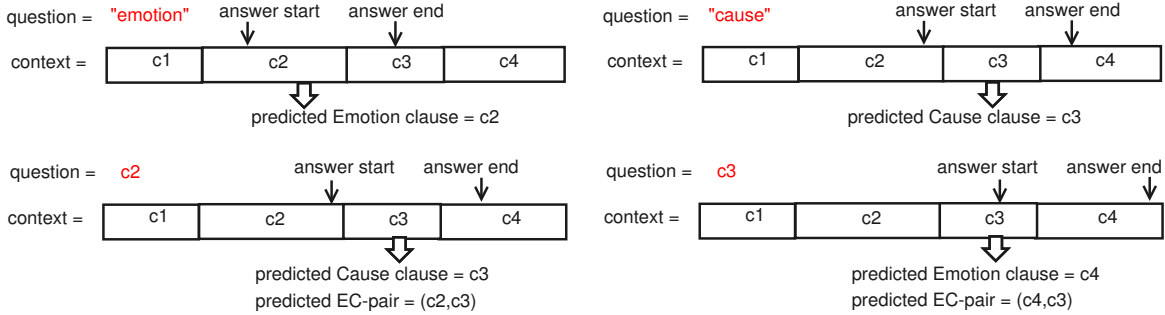, height=1.65in}
\caption{Guided pair extraction Guided-QA: Emotion is detected first (Left), Cause is detected first (Right).}
\label{fig:guided-qa}
\end{figure*}

% In this section, we describe our QA-based method for the ECPE task. We start with the baseline method named Indep-QA which tries to predict emotion clause and cause clause independently. Then we propose Guided-QA scheme to better model the relationship between emotion and cause clauses.

\subsection{Problem Statement}
%We followed the formal definition of the ECPE stated in \cite{xia2019emotion}. Briefly, g

Given a document of $n$ clauses $d = (c_1,c_2,..,c_n)$, the goal of ECPE is to detect all potential emotion-cause pairs $P=\{..(c_e, c_c),..\}$ where $c_e$ is an emotion clause, and $c_c$ is the corresponding cause clause \cite{xia2019emotion}. We formulated the ECPE task as a QA problem. Given a set of questions $\{q_e, q_c\}$ ($q_e$ is for emotion and $q_c$ is for cause) and a context document $d$ with $n$ clauses, the model learns to predict start and end positions of each $c_e$ and $c_c$: $s_{c_e}, e_{c_e} = f(d, q_e|\Theta)$ and $s_{c_c}, e_{c_c} = f(d, q_c|\Theta)$ to form $P$. $\Theta$ can be learnt by using independent or guided extraction.

%the task is to extract relevant emotion-cause pairs $(c_e, c_c)$ to form $P$. The extraction can be done by predicting the start and end positions of each $c_e$ and $c_c$ as $s_{c_e}, e_{c_e} = f(d, q_e|\Theta)$ and $s_{c_c}, e_{c_c} = f(d, q_e|\Theta)$. $\Theta$ can be learnt by using independent or guided extraction as follows.

\subsection{Independent Emotion, Cause Extraction}
We first introduce a simple version of our model, Indep-QA in Figure \ref{fig:indep-qa}.
Indep-QA receives a fixed question (for emotion or cause) and then pulls out corresponding emotion or cause clauses independently.

%outlines the basic idea of our Indep-QA. Emotion clauses and cause clauses are detected independently with generic questions. 
% Given a document, we first concatenate all the clauses to a long context \textit{C}. We use fixed questions, "emotion" for emotion and "cause" for cause. This means each test document will have one predicted span for emotion clause and one for cause clause. 

\paragraph{Question formulation}
Because no emotion/cause information is provided beforehand, we have to detect them first with generic questions. It is possible to use pre-defined questions for extraction \cite{mengge2020coarse}, however, we argue that the definition of questions is time-consuming, needs domain knowledge, and does not guarantee the semantic relationship between the questions and context documents. Instead, we use two short questions "\texttt{emotion}" and "\texttt{cause}" as an implicit indicator that provides additional information for the model. We leave the analysis of using generic questions such as "What is the emotion?" and "What is the cause?" as future work.

\paragraph{Learning and prediction}
Given a document $d$ and a question ("\texttt{emotion}" or "\texttt{cause}"), we concatenated all clauses of $d$ and the question to form a single sequence $C$. The sequence was fed to a pretrained language model (PLM) to obtain its hidden representations of tokens which were subsequently fed into a feed-forward layer followed by a softmax layer. Each candidate span was scored as the product of start/end probabilities. The maximum scoring span was used as the prediction.
% We used the context vector from the [CLS] token for the prediction of start and end positions.

\paragraph{Mapping predicted answer span to clauses}
The predicted answer span may overlap with one or several clauses. We applied a span-to-clause mapping rule to determine which clauses are predicted results: the clause that overlaps most with the predicted span is returned. The tie is broken arbitrarily. For instance, In Figure \ref{fig:indep-qa}, the predicted span for "emotion" overlaps with clauses $c2$ and $c3$ in which $c_2$ is more overlapped. As a result, $c_2$ is the predicted emotion.

\paragraph{EC pair prediction} 
% The predicted emotion-cause pair of each document is ($c_e$, $c_c$) where $c_e$, $c_c$ are predicted emotion/cause clauses respectively.
Given predicted emotion/cause clauses $c_e$ and $c_c$, Indep-QA simply predicts ($c_e$, $c_c$) as an emotion-cause pair. As illustrated in Figure \ref{fig:indep-qa}, $(c_2, c4)$ is the predicted emotion-cause pair.

\subsection{Guided Emotion-Cause Pair Extraction}
The Indep-QA model extracts emotion/clause clauses independently but does not exploit the relationship between emotion and cause clauses, which plays an important role in the extraction of emotion-cause pairs \cite{ding2020end,wei2020effective,chen2020end,cheng2021unified,chen2022recurrent}. 

To better model this relationship, we introduce Guided-QA in Figure \ref{fig:guided-qa}. The model receives an emotion question and predicts the corresponding emotion clause. Then the predicted emotion clause is used as a question for cause extraction. Compared to Indep-QA, the Guided-QA takes into account an implicit relationship from emotion for cause prediction.

The Guided-QA model shares the question formulation, hidden representation learning, and the mapping process of the Indep-QA model.

% \paragraph{Emotion prediction}
% The model starts with emotion prediction in the same way as Indep-QA, i.e. the generic question "emotion" was used.

% As in Indep-QA, we first predict emotion clause using a fixed question "emotion", say $c_2$. 
\paragraph{EC pair extraction}
We used the predicted (noisy) emotion clause as the question for cause extraction. The interaction between emotion and cause happens here. The predicted emotion clause may or may not be the true one but on average, it contains much more information for the QA model than the generic question (i.e., "emotion"). 
%Again, other choices of question formulation for cause extraction such as building questions using keywords in the predicted emotion are left for future work. 
Note that the predicted (noisy) emotion as the question was used for the test set only. For the training set, as the model already knows which clauses are emotion or cause, it uses the true emotion clause as the question. 

By swapping the role, the model can detect cause clauses first and use the noisy causes as questions to predict the emotions. 
In Section \ref{sec:result} we compare Emotion-first and Cause-first, the two variants of Guided-QA and show that the gaps are tiny. In other word, the two variants are almost equivalent on the tested datasets.
% We only report the results of using predicted emotion for cause prediction due to tiny gaps between the two methods.

As our QA models use the best answer span for each question, only one emotion, one cause, and one EC pair are predicted for each document which are appropriate for the ECPE dataset. We also aware that the prediction of spans should be multiple and we aim to address this limitation in future work by using multiple span extraction methods \cite{nguyen2021span,fu2021spanner}.

% We applied the same span-to-clause mapping rule as in Indep-QA. Each document accepts one emotion-cause pair ($c_e$, $c_c$) where $c_c$ is the answer of the question $c_e$. 

\subsection{Discussion}
Given a document of $n$ clauses, existing schemes such as ECPE-MLL \cite{ding2020end}, RankCP \cite{wei2020effective} and PairGCN \cite{chen2020end} attempt to reduce the $O(n^2)$ complexity of emotion-cause pair classification by using sliding window, transition graph techniques. However, these techniques may miss certain interaction between the emotion-cause pair and the full context in the document. BERT-based QA models with full attention between the question and the context mitigate this issue. Through QA models, the emotion-cause relationship between all clauses is implicitly learned and we can leverage the power of existing QA methods.

\section{Experimental Settings}

\begin{table*}[!t]
\centering
\caption{Histogram of the number of emotion-cause pairs per document.} \label{tab:data-stats}
\begin{tabular}{|c|c|c|}
\hline
 & Number & Percentage \\
\hline
Documents with one emotion-cause pair & 1746 & 89.77\% \\
Documents with two emotion-cause pairs & 177 & 9.10\% \\
Documents with more than two emotion-cause pairs & 22 & 1.13\% \\
All & 1945 & 100\% \\ 
\hline
\end{tabular}%\vspace{-0.2cm}
\end{table*}

\begin{table*}[t]
\centering
\caption{Guided-QA Emotion-first vs. Cause-first on 10-split ECPE dataset and 20-split TransECPE dataset} 
\label{tab:result-guided-qa}
\begin{tabular}{l ccc ccc ccc} \hline
 \textbf{Model} & \multicolumn{3}{c}{Emotion Extraction} & \multicolumn{3}{c}{Cause Extraction} & \multicolumn{3}{c}{EC Pair Extraction} \\ \cline{2-10}
  & P & R & F1 &  P & R & F1 &  P & R & F1 \\ 
\hline
    \textbf{10-split ECPE} & & & & & & & & & \\
% \hline
    Emotion-first (BERT) & 0.847 & 0.908 & 0.876 & 0.719 & 0.792 & 0.754 & 0.771 & 0.692 & 0.729 \\
	Cause-first (BERT) & 0.831 & 0.891 & 0.860 & 0.714 & 0.787 & 0.749 & 0.763 & 0.685 & 0.722\\
% \hline
	Emotion-first (RoBERTa) & 0.854 & 0.916 & 0.884 & 0.732 & 0.806 & 0.767 & 0.786 & 0.706 & 0.744  \\
	Cause-first (RoBERTa) & 0.843 & 0.904 & 0.873 & 0.733 & 0.807 & 0.768 & 0.784 & 0.704 & 0.742 \\ 
\hline
    \textbf{20-split TransECPE} & & & & & & & & & \\    
% \hline
    Emotion-first (BERT) & 0.842 & 0.906 & 0.873 & 0.710 & 0.782 & 0.744 & 0.760 & 0.689 & 0.723 \\ 
	Cause-first (BERT) & 0.833 & 0.897 & 0.864 & 0.713 & 0.785 & 0.747 & 0.761 & 0.690 & 0.724 \\
% \hline
	Emotion-first (RoBERTa) & 0.844 & 0.909 & 0.875 & 0.723 & 0.796 & 0.757 & 0.772 & 0.700 & 0.734 \\ 
	Cause-first (RoBERTa) & 0.838 & 0.902 & 0.869 & 0.724 & 0.797 & 0.758 & 0.773 & 0.701 & 0.735 \\ 
\hline	
\end{tabular}
\end{table*}

\begin{table*}[t]
\centering
\caption{Experimental results of different models on 10-split ECPE dataset. * indicates reproduced results.} 
% * denotes our rerun results. \textbf{Bold} values are significant with $p \leq 0.05$ compared to Indep-QA and rerun methods.
\label{tab:result-ecpe}
\begin{tabular}{l ccc ccc ccc} \hline
 \textbf{Model} & \multicolumn{3}{c}{Emotion Extraction} & \multicolumn{3}{c}{Cause Extraction} & \multicolumn{3}{c}{EC Pair Extraction} \\ \cline{2-10}
  & P & R & F1 &  P & R & F1 &  P & R & F1 \\ \hline
  
    Indep-QA (BERT) & 0.847 & 0.908 & \textbf{0.876} & 0.714 & 0.787 & 0.749 & 0.736 & 0.661 & 0.697 \\
	Guided-QA (BERT) & 0.847 & 0.908 & \textbf{0.876} & 0.719 & 0.792 & \textbf{0.754} & 0.771 & 0.692 & \textbf{0.729} \\
\hline
	Indep-QA (RoBERTa) & 0.854 & 0.916 & 0.884 & 0.733 & 0.807 & 0.768 & 0.761 & 0.683 & 0.720 \\
	Guided-QA (RoBERTa) & 0.854 & 0.916 & 0.884 & 0.732 & 0.806 & 0.767 & 0.786 & 0.706 & \textit{\textbf{0.744}} \\ \hline \hline
    
	ECPE-MLL (BERT) & 0.861 & 0.919 & 0.889 & 0.738 & 0.791 & 0.763 & 0.770 & 0.724 & 0.745 \\
	RankCP (BERT) & 0.912 & 0.900 & 0.906 & 0.746 & 0.779 & 0.762 & 0.712 & 0.763 & 0.736 \\
	PairGCN (BERT) & 0.886 & 0.796 & 0.838 & 0.791 & 0.693 & 0.738 & 0.769 & 0.679 & 0.720 \\ 
	UTOS (BERT) & 0.882 & 0.832 & 0.856 & 0.767 & 0.732 & 0.747 & 0.739 & 0.706 & 0.720 \\ 
	RSN (BERT) & 0.861 & 0.892 & 0.876 & 0.773 & 0.740 & 0.755 & 0.760 & 0.722 & 0.739 \\ \hline
	ECPE-MLL (BERT)* & --- & --- & --- & --- & --- & --- & 0.688 & 0.752 & 0.718 \\
	RankCP (BERT)* &  0.741 & 0.744 & 0.742 & 0.614 & 0.647 & 0.627 & 0.573 & 0.625 & 0.597 \\
	PairGCN (BERT)* & 0.784 & 0.883 & 0.829 & 0.686 & 0.795 & 0.735 & 0.675 & 0.772 & 0.718 \\ \hline
\end{tabular}
\end{table*}

\begin{table*}[t]
\centering
\caption{Experimental results of different models on 20-split TransECPE dataset. * indicates reproduced results. The authors of PairGCN and RSN did not tested their models on TransECPE.} 
% * denotes our rerun results. 
\label{tab:result-transecpe}
\begin{tabular}{l ccc ccc ccc} \hline
 \textbf{Model} & \multicolumn{3}{c}{Emotion Extraction} & \multicolumn{3}{c}{Cause Extraction} & \multicolumn{3}{c}{EC Pair Extraction} \\ \cline{2-10}
 & P & R & F1 &  P & R & F1 &  P & R & F1 \\ \hline

	Indep-QA (BERT) & 0.842 & 0.906 & 0.873 & 0.713 & 0.785 & \textbf{0.747} & 0.730 & 0.662 & 0.694  \\
	Guided-QA (BERT) & 0.842 & 0.906 & 0.873 & 0.710 & 0.782 & 0.744 & 0.760 & 0.689 & \textbf{0.723} \\ \hline
	Indep-QA (RoBERTa) & 0.844 & 0.909 & 0.875 & 0.724 & 0.797 & 0.758 & 0.739 & 0.670 & 0.703 \\ 
	Guided-QA (RoBERTa) & 0.844 & 0.909 & 0.875 & 0.723 & 0.796 & 0.757 & 0.772 & 0.700 & \textit{\textbf{0.734}} \\ \hline \hline
    
	ECPE-MLL (BERT) & 0.847 & 0.899 & 0.872 & 0.705 & 0.770 & 0.736 & 0.749 & 0.698 & 0.722\\
	RankCP (BERT) & 0.894 & 0.895 & 0.894 & 0.694 & 0.747 & 0.719 & 0.658 & 0.731 & 0.692 \\ 
	UTOS (BERT) & 0.865 & 0.829 & 0.849 & 0.742 & 0.708 & 0.728 & 0.710 & 0.681 & 0.691 \\ \hline
	ECPE-MLL (BERT)* & --- & --- & --- & --- & --- & --- & 0.659 & 0.714 & 0.684 \\
	RankCP (BERT)* & 0.896 & 0.897 & \textbf{0.896} & 0.694 & 0.749 & 0.720 & 0.657 & 0.731 & 0.691 \\
	PairGCN (BERT)* & 0.804 & 0.878 & 0.839 & 0.689 & 0.770 & 0.727 & 0.677 & 0.746 & 0.709 \\ \hline
	
\end{tabular}%\vspace{-0.2cm}
\end{table*}

\paragraph{Datasets}
We followed the 10-split ECPE dataset provided by \cite{xia2019emotion} and the 20-split TransECPE variant \cite{fan2020transition} to evaluate our methods.  Each split is a random partition of the 1945 documents to train/dev/test sets with ratio 8:1:1, i.e., the train set, dev set and test set contain approximately 1556, 194 and 195 documents. On average, each document contains 14.8 clauses.

Table \ref{tab:data-stats} shows the distribution of documents with different number of emotion-cause pairs. Most of the documents have only one emotion-cause pairs. This fact makes the detection of emotion/cause clauses as well as emotion-cause pairs challenging.

\paragraph{Evaluation metrics}
We used the precision, recall, and F1 score \cite{xia2019emotion} as evaluation metrics for all three tasks of ECPE: emotion extraction, cause extraction and emotion-cause pair extraction. Let $T_e$ and $P_e$ be the number of ground-truth and predicted emotion clauses respectively, the precision, recall and F1 score for emotion are as defined as follows.

\[
P_e = \frac{| T_e \cap P_e |}{| P_e |} 
\]

\[
R_e = \frac{| T_e \cap P_e |}{| T_e |} 
\]

\[
F1_e = \frac{2*P_e*R_e}{P_e + R_e}
\]

Metrics for cause clauses and emotion-cause pairs are defined similarly.

\paragraph{Implementation details}
Our model was implemented using BERT classes provided by Huggingface \cite{wolf2020transformers}. The model was trained in 5 epochs, with the learning rate of $5e-5$, and the batch size of 16. We used BERT \cite{devlin2018bert}\footnote{https://huggingface.co/bert-base-chinese}  and RoBERTa  \cite{liu2019roberta}\footnote{https://huggingface.co/hfl/chinese-roberta-wwm-ext} for Chinese. All models were trained on a Tesla P100 GPU.

%\paragraph{Baselines}
%We compare our model to five strong methods for ECPE, namely ECPE-MLL\footnote{https://github.com/NUSTM/ECPE-MLL} \cite{ding2020end}, RankCP\footnote{https://github.com/Determined22/Rank-Emotion-Cause} \cite{wei2020effective}, PairGCN\footnote{https://github.com/chenying3176/PairGCN\_ECPE} \cite{chen2020end}, UTOS \cite{cheng2021unified}, and RSN \cite{chen2022recurrent}. For fair comparison \cite{houghton2020guaranteeing}, we also rerun the each publicly available source code in the original setting.
% renard2020variability

\section{Results and Discussion}
\label{sec:result}
\paragraph{Guided-QA: Emotion-first vs. Cause-first}
We first compare the two variants Emotion-first and Cause-first of the Guided-QA method. Table \ref{tab:result-guided-qa} shows that the two variants have almost equivalent performance on the tested datasets except the BERT-based results on 10-split ECPE. Also, the RoBERTa-based results are consistently better than the BERT-based, 1.1 to 2.0 points.
In the next section, we pick the Emotion-first scores for comparing Guided-QA with other methods.

\paragraph{Guided-QA vs. Indep-QA}
We now compare Guided-QA and Indep-QA. For 10-split ECPE in the upper part of Table \ref{tab:result-ecpe}, the Guided-QA model is consistently better than Indep-QA for pair extraction. This is because Guided-QA takes into account the implicit interaction between emotion and cause clauses. For emotion or cause extraction, Indep-QA is competitive with Guided-QA. This is because they share the same formulation. The results in Table \ref{tab:result-transecpe} also show similar observation.

We also confirm the performance of our model by using RoBERTa to have better analysis. The results are consistent with the model using BERT, in which Guided-QA outputs better F-scores than the Indep-QA model. It also shows that our model can be improved further by using stronger PLMs.

\paragraph{Guided-QA vs. strong baselines}
We compare our model with five strong methods for ECPE: ECPE-MLL\footnote{https://github.com/NUSTM/ECPE-MLL} \cite{ding2020end}, RankCP\footnote{https://github.com/Determined22/Rank-Emotion-Cause} \cite{wei2020effective}, PairGCN\footnote{https://github.com/chenying3176/PairGCN\_ECPE} \cite{chen2020end}, UTOS \cite{cheng2021unified}, and RSN \cite{chen2022recurrent}. For 10-split, our model using BERT follows ECPE-MLL, RankCP, and RSN. It shows that with a simple architecture, our model can output competitive results compared to complicated methods. For 20-split TransECPE in Table \ref{tab:result-transecpe}, the trend is consistent with Table \ref{tab:result-ecpe}, in which the Guided-QA model is competitive for both ECE and ECPE tasks.

Moreover, as we observe from all the compared methods, the gaps between the reported pair-f1 scores for 10-split ECPE and 20-split TransECPE are 0.023 (=0.745-0.722) for ECPE-MLL, 0.042 for RankCP, 0.029 for UTOS, 0.003 for Indep-QA and 0.006 for Guided-QA, i.e., largest gap in RankCP and smallest gaps in our models. Across the two settings, our models seem more robust than the compared methods.

%Also note that the results from the original papers are just for reference because it seems there are gaps between the reproduced results and original results.\footnote{https://github.com/Determined22/Rank-Emotion-Cause/issues/3}

\paragraph{Reproducibility}
For fair comparison \cite{houghton2020guaranteeing}, we also rerun  publicly available source codes in the original setting. The reproduced results confirm the gaps between reproduction and original results. Compared to the reproduced results, Guided-QA using BERT is the best for EC pair extraction.

%% OLD content
Compared to the results of reproduced methods, the Guided-QA is still better for both ECE and ECPE tasks. This confirms our hypotheses stated in Section \ref{sec:intro}. Compared to the results of strong baselines reported in papers, the F-scores of Guided-QA are still competitive. It shows that our simple model can output promising results compared to complicated ECPE methods \cite{ding2020end,wei2020effective,chen2020end,cheng2021unified,chen2022recurrent}. The results from the original papers are just for reference because it seems there are gaps between the reproduced results and original results.\footnote{https://github.com/Determined22/Rank-Emotion-Cause/issues/3}
. This is because several scholars tried to reproduce the results, but it seems there are gaps between the reproduced results and original results.

For 20-split TransECPE in Table \ref{tab:result-transecpe}, the trend is consistent with Table \ref{tab:result-ecpe}. The Guided-QA is competitive for both ECE and ECPE tasks. The model using RoBERTa is still the best. After rerunning the source codes of the baselines, we found that PairGCN has the best reproducibility.

By adopting the standardized pipeline of BERT-based question answering, our models inherit its simplicity and reproducibility which may become an issue in more complex methods like RankCP.

%% OLD content
\paragraph{Runtime comparison}
We also measured the running time of our model and the baselines. In Table \ref{tab:runtime}, PairGCN which only uses BERT embeddings has the best running time. The other models take longer to run due to the fine-tuning of BERT models. Our model is the second best, which is much faster than ECPE-MLL. It shows that our model can balance between competitive accuracy and high speed.

\begin{table}[!t]
\centering
\caption{Running time (train and test) on Tesla P100.} \label{tab:runtime}
%\begin{tabular}{|l|r|r|r|r|p{12mm}|p{12mm}|}
\begin{tabular}{|l|rr|}
\hline
 & ECPE & TransECPE \\
\hline
ECPE-MLL & 8.5h & 17h \\
RankCP &  3h & 6h \\
PairGCN & 42min & 85 min \\
\hline
Indep-QA & 2h30 & 5h \\
Guided-QA & 2h30 & 5h \\
\hline
\end{tabular}%\vspace{-0.2cm}
\end{table}

\section {Conclusion}
This paper introduces a paradigm shift for the ECPE task. Instead of treating the task as the conventional formulation, we formulate the extraction as a QA problem. Based on that, we design a model which takes into account the implicit interaction between emotion and cause clauses. Experimental results on a benchmark Chinese dataset show that using implicit interaction of emotions and causes can achieve competitive accuracy compared to strong baselines. Future work will consider explicit interaction between emotion and cause clauses.

%ECPE is a challenging task because we have to extract emotions, causes and EC pairs simultaneously. We re-emphasize that pair extraction task is much more important than independent emotion/cause extraction.
%With Guided-QA, we have shown that the standard BERT-based extractive QA models are effective for the ECPE task. Standard BERT-based QA models are not only easy to setup but also give better reproducibility. 

%ECPE is a challenging task because we have to extract emotions, causes and EC pairs simultaneously. We re-emphasize that pair extraction task is much more important than independent emotion/cause extraction.
%With Guided-QA, we have shown that the standard BERT-based extractive QA models are effective for the ECPE task. Standard BERT-based QA models are not only easy to setup but also give better reproducibility. 

% Entries for the entire Anthology, followed by custom entries
\bibliography{anthology,custom}
\bibliographystyle{acl_natbib}

% \appendix

% \section{Example Appendix}
% \label{sec:appendix}

% This is an appendix.

\end{document}